\documentclass[sigconf]{acmart}

\usepackage{graphicx}
\usepackage[colorinlistoftodos]{todonotes}
\usepackage{multirow}

\usepackage[ruled,vlined]{algorithm2e}

\SetKwFor{For}{for (}{) $\lbrace$}{$\rbrace$}

\theoremstyle{remark}

\setcopyright{none} 
\acmConference[Anonymous Submission to ACM CCS 2020]{ACM Conference on Computer and Communications Security}{Due 15 May 2019}{London, TBD}
\acmYear{2020}

\begin{document}
\title{Privacy-preserving Transfer Learning via Secure Maximum Mean Discrepancy} 
\author{Bin Zhang}
\affiliation{\small{Tongji University}}

\author{Cen Chen}
\affiliation{\small{Alibaba Group}}

\author{Li Wang}
\affiliation{\small{Alibaba Group}}

\begin{abstract}
The success of machine learning algorithms often relies on
a large amount of high-quality data to train well-performed models. 
However, data is a valuable resource and are always held by different parties in reality. An effective solution to such a ``data isolation" problem is to employ federated learning, which allows multiple parties to collaboratively train a model. 
In this paper, we propose a Secure version of the widely used Maximum Mean Discrepancy (SMMD) based on homomorphic encryption to enable effective knowledge transfer 
under the data federation setting without compromising the data privacy.
The proposed SMMD is able to avoid the potential information leakage in transfer learning when aligning the source and target data distribution. As a result, both the source domain and target domain can fully utilize their data to build more scalable models. Experimental results demonstrate that our proposed SMMD is secure and effective.

\end{abstract}



\maketitle

\section{Introduction}
Witness the rapid development and significant success in various applications, machine learning is gradually becoming a powerful production tool for many organizations in recent years thanks to the availability of large scale datasets \cite{ImageNet, Cityscapes}. Numerous deep learning models \cite{ResNet} were born and achieved state-of-the-art performance on those prevalent benchmarks. However, annotating high-quality labeled data is a very time-consuming process, especially for those dense prediction tasks such as semantic segmentation. Transfer learning \cite{Transfer} is an effective way to solve the difficulty of data annotation by transferring knowledge from a related source domain to the target domain. Besides, data required for training the models may not be owned by a single organization and is always stored across different institutions. Due to business competition, data privacy issues, and regulatory requirements, different companies often cannot share their own user data. Therefore, how to jointly train machine learning models without leaking individual data privacy is of great significance for deploying machine learning models to practical applications.

As companies pay more and more attention to privacy protection, data privacy has become a topic of common concern all over the world. In recent years, there has been a lot of news about the abuse of data by the government and various companies. For example, Facebook's massive user data leakage event caused strong repercussions. In order to overcome those challenges, Google made
the first attempt to introduce a federated learning framework \cite{Federated}, in which several distributed parties train a machine learning model collaboratively without exposing their raw data. In federated learning, encrypted gradients are transmitted between independent organizations and thus data owners may not suffer from privacy leakage. Federated learning can be divided into two categories, i.e., horizontal federated learning and vertical federated learning. In horizontal federated learning, data records share the same feature space but differ in the sample space. For example, considering e-commerce companies located in different countries, their users are from different regions but their business is similar, thus their feature spaces may be identical. Similarly, vertical federated learning refers to the situation where sample space is the same but feature space is different. In this paper, we focus on the more challenging federated learning setting where \textit{datasets may differ in both the sample and feature space}. 
In such a setting, we need to deal with heterogeneous data sources and transfer the knowledge safely without violating the privacy protection protocol.
\begin{figure}[t!]
\centerline{\includegraphics[width=240pt,height=135pt]{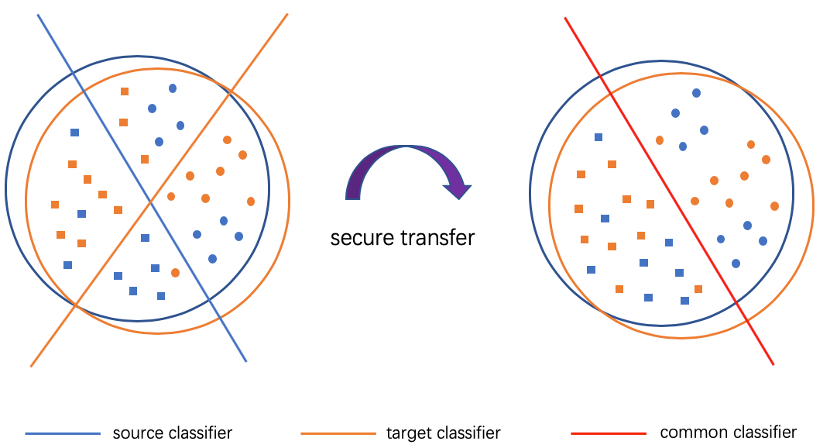}}
\caption{Illustration of secure transfer learning}
\label{fig1}
\end{figure}

To address the aforementioned challenges, inspired by the success of transfer learning on effectively leveraging multiple data sources, we propose a novel technique called Secure Maximum Mean Discrepancy (SMMD) to prevent the potential privacy leakage when adopting the widely used maximum mean discrepancy (MMD) alignment loss for knowledge transfer~\cite{MMD}. We incorporate Homomorphic Encryption (HE) \cite{rivest1978data} into existing maximum mean discrepancy loss computation and thus help transfer information from the source domain to the target domain safely. We conduct extensive experiments on various datasets and the results demonstrate that our proposed SMMD is secure and efficient. To the best of our knowledge, we are the first to introduce a privacy protection mechanism to standard MMD and our SMMD can be adapted to many transfer learning settings.

\section{Related work}
\subsection{Federated Learning}
Federated learning \cite{FedApp} is a machine learning framework that can effectively help multiple institutions to perform data usage and machine learning modeling under the requirements of user privacy protection, data security, and government regulations. As a distributed machine learning paradigm, federated learning can effectively solve the problem of data islands, allowing participants to jointly model on the basis of not sharing data, which can technically break the data islands and achieve AI collaboration. In \cite{DPFederated}, they proposed a technique that introduces differential privacy into federated learning to protect the client data. Like machine learning, federated learning also requires that the training data is independently and identically distributed, and the performance of federated learning will be greatly reduced when faced with non-independently and identically distributed data. In \cite{NonIID} they introduced an effective mechanism to deal with the Non-IID data. Under normal circumstances, the number of devices in federated learning is often large, and different devices may differ in the data size, data feature distribution, and available resources. In order to solve the above problems, \cite{li2019fair} proposed an optimization algorithm to ensure that the performance of the federated learning model is fairly distributed among various devices.

\subsection{Transfer Learning}
Transfer learning provides an effective approach to solving the domain shift \cite{unbias} problem between different datasets and enables knowledge to transfer across domains. One of the most critical issues in transfer learning is how to measure the distance between the data distribution of the source and target domains and numerous approaches have been proposed \cite{MMD, kifer2004detecting, koltchinskii2001rademacher, mansour2009domain}. By adding an adaptation layer between the source domain and the target domain and adding a domain confusion loss function to allow the network to learn how to classify, DDC \cite{tzeng2014deep} reduced the distribution difference between the source domain and the target domain and thus achieved knowledge transfer. However, DDC only fits one layer of the network and uses only a single kernel MMD loss function, which might not be optimal. To overcome these issues, DAN \cite{long2015learning} proposes a new deep adaptive network structure, which uses the multi-kernel optimization selection method of mean embedded matching to further reduce the domain gap. Inspired by the generative adversarial network (GAN) \cite{goodfellow2014generative}, \cite{shrivastava2017learning} incorporated adversarial training into the domain adaptation process where a domain classifier was applied to achieve domain-invariant feature extraction. In \cite{liu2018secure}, they provided a novel secure federated transfer learning framework that equips conventional neural networks with additively homomorphic encryption (HE) and multi-party computation (MPC). Unlike existing transfer learning methods that deal with the homogeneous feature space, \cite{gao2019privacy} introduced a secure multi-party computation protocol to mitigate the domain shift in heterogeneous feature space.

\section{Problem Definition}
Consider a labeled source domain dataset $X^{s}=\left\{\left(\mathbf{x}_{i}^{s}, y_{i}^{s}\right)\right\}_{i=1}^{N_{s}}$
with $N_{s}$ annotated samples, where $\mathbf{x}_{i}^{s}$ is the i-th sample drawn from a source distribution $P_{S}(\mathbf{x},y)$.  $y_{i}^{s}$ is the corresponding label. Analogously, we represent the unannotated target domain as $X^{t}=\left\{\mathbf{x}_{i}^{t}\right\}_{i=1}^{N_{t}}$, where $\mathbf{x}_{i}^{t}$ is the i-th sample drawn from a target distribution $P_{T}(\mathbf{x},y)$, and $N_{t}$ is the total sample number in target domain. 
We also consider the existence of a set of co-occurrence data samples $X^{st}=\left\{\left(\mathbf{x}_{i}^{s}, x_{i}^{t}\right)\right\}_{i=1}^{N_{st}}$ .We focus on the transfer learning setting so we assume that the source and target domain have different data distribution, i.e., $P_{S}(\mathbf{x},y)\not=P_{T}(\mathbf{x},y)$. As illustrated in Figure.1,
we aim to design a classifier $y=f_{s}(\mathbf{x})$ and transfer it to the target domain without information leakage, such that the expected target risk $R_{t}=\mathbb{E}_{(\mathbf{x}, y) \sim P_{T}}\left[f_{s}(\mathbf{x}) \neq y\right]$ is bounded by utilizing the well-annotated source domain data.
\begin{figure}[t!]
\centerline{\includegraphics[width=240pt,height=135pt]{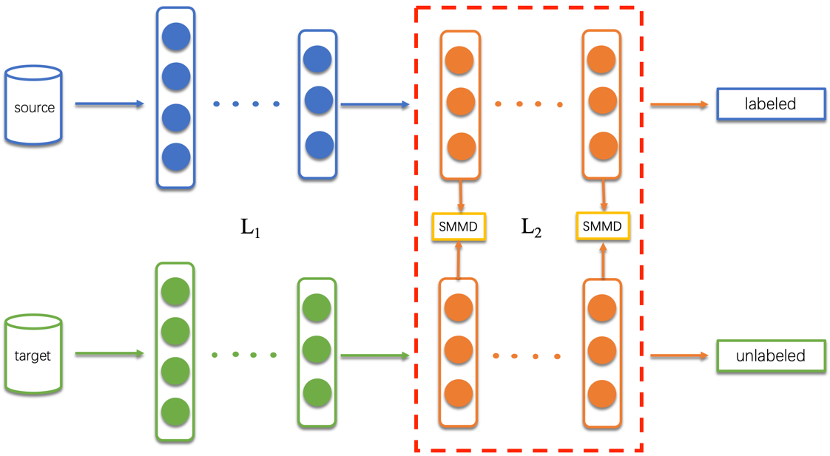}}
\caption{Overall framework of our proposed SMMD}
\label{fig1}
\end{figure}

\section{Proposed Method}
\subsection{Introduction to MMD}
Maximum Mean Discrepancy (MMD) is a widely used method to measure the distance between the data distribution of source domain and target domain. Given the above source domain dataset $X^{s}$ and target domain dataset $X^{t}$, MMD is defined as the mean embedding difference between the two sets of samples:
\begin{equation}
    \small
	\begin{aligned}
		& \mathrm{MMD}^{2}[X^{s}, X^{t}] \\
		=&\left\|\mathbf{E}_{x^{s}\sim \operatorname{P_{S}(\mathbf{x},y)}}[\phi(\mathbf{x^{s}})]-\mathbf{E}_{x^{t}\sim \operatorname{P_{T}(\mathbf{x},y)}}[\phi(\mathbf{x^{t}})]\right\|^{2} \\
		=&\left\|\frac{1}{N_{S}} \sum_{i=1}^{N_{S}} \phi\left(\mathbf{x}^{s}_{i}\right)-\frac{1}{N_{T}} \sum_{j=1}^{N_{T}} \phi\left(\mathbf{x}^{t}_{j}\right)\right\|^{2} \\
		=& \frac{1}{N_{S}^{2}} \sum_{i=1}^{N_{S}} \sum_{i^{\prime}=1}^{N_{S}} \phi\left(\mathbf{x}^{s}_{i}\right)^{T} \phi\left(\mathbf{x}^{s}_{i^{\prime}}\right)+\frac{1}{N_{T}^{2}} \sum_{j=1}^{N_{T}} \sum_{j^{\prime}=1}^{N_{T}} \phi\left(\mathbf{x}^{t}_{j}\right)^{T} \phi\left(\mathbf{x}^{t}_{j^{\prime}}\right) \\
		&-\frac{2}{N_{S} N_{T}} \sum_{i=1}^{N_{S}} \sum_{j=1}^{N_{T}} \phi\left(\mathbf{x}^{s}_{i}\right)^{T} \phi\left(\mathbf{x}^{t}_{j}\right),
	\end{aligned}
\end{equation}
where $\phi(\cdot)$ is the feature mapping function.

\subsection{Secure MMD}
Formally, consider a source domain network $C_{s}$ and a target domain network $C_{t}$, which take the source domain data ${x}_{i}^{s}$ and target domain data ${x}_{i}^{t}$ as input, respectively. Suppose there are total $L$ layers in each network, where $L=L_{1}+L_{2}$. We adopt a weakly-shared transfer learning manner and align the later $L_{2}$ layers of source and target network, as shown in Figure.2. Let $h_{l}\left(x_{s}\right)$ and $h_{l}\left(x_{t}\right)$ denote the hidden representation of layer $l$ extracted by the source domain and target domain network, respectively. Following \cite{shu2015weakly}, we also use a translator function to provide pseudo label for the co-occurrence unlabeled target data. The classification objective using the co-occurrence set can be summarized as:
\begin{equation}
\small
\underset{\Theta^{s}, \Theta^{t}}{\operatorname{argmin}} \mathcal{L}_{cls}=\sum_{i}^{N_{st}} \ell_{1}\left(y_{i}^{s}, y_{i}^{t}\right),
\end{equation}
where $\Theta^{s}$ and $\Theta^{t}$ are the network parameter of source and target domain network, $y_{i}^{t}$ is the pseudo label of target data provided by the available source data. We adopt the conventional logistic loss for the classification task. 

In addition to the logistic loss $\ell_{1}$ for classification, we also need to align the intermediate hidden representation of source and target domain data. The first $L_{1}$ layers of source and target network are expected to learn domain-specific feature while the latter $L_{2}$ layers are supposed to learn domain-invariant feature. We minimize the following loss function to achieve this goal:
\begin{equation}
\small
\underset{\Theta^{s}, \Theta^{t}}{\operatorname{argmin}} \mathcal{L}_{mmd}=\sum_{i=L_{1}+1}^{L_{2}} \ell_{2}\left(h_{i}\left(x_{s}\right), h_{i}\left(x_{t}\right)\right),
\end{equation}
where $L_{2}$ is the alignment loss, we choose the widely used MMD as our alignment function. In order to ensure there is no information leakage in the feature alignment process, we add additively homomorphic encryption to the original MMD loss.
\subsection{Training Objective and Optimization}
By combining the aforementioned loss functions, we can summarize the overall optimization objective as:
\begin{equation}
\small
\underset{\theta^{s}, \theta^{t}}{\operatorname{argmin}} \mathcal{L}=\mathcal{L}_{cls}+\alpha \mathcal{L}_{mmd}+\frac{\beta}{2}\left(\mathcal{L}_{reg}^{s}+\mathcal{L}_{reg}^{t}\right)
\end{equation}
where $\mathcal{L}_{reg}^{s}$ and $\mathcal{L}_{reg}^{t}$ are the regularization term for source and target network. For the classification loss, we use the second order Taylor expansion to calculate the gradient during the backpropagation process. Applying equation (4) with additively homomorphic encryption $[[\cdot]]$:
\begin{equation}
\small
\begin{aligned}
\mathcal{L} &=\sum_{i}^{N_{st}}\left(\left[\left[\ell_{1}\left(y_{i}^{s}, y_{i}^{t}\right)\right]\right]\right) \\
&+ \frac{\alpha}{N_{st}^{2}} \sum_{i=1}^{N_{st}} \sum_{j=1}^{N_{st}} \left[\left[k\left(h_{i}^{s}, h_{j}^{s}\right)\right]\right]+\frac{\alpha}{N_{st}^{2}} \sum_{i=1}^{N_{st}} \sum_{j=1}^{N_{st}} \left[\left[k\left(h_{i}^{t}, h_{j}^{t}\right)\right]\right] \\
&-\frac{2\alpha}{N_{st}^{2}} \sum_{i=1}^{N_{st}} \sum_{j=1}^{N_{st}}\left[\left[k\left(h_{i}^{s}, h_{j}^{t}\right)\right]\right]
+\left[\left[\frac{\beta}{2} \mathcal{L}_{reg}^{s}\right]\right]+\left[\left[\frac{\beta}{2} \mathcal{L}_{reg}^{t}\right]\right] \\
&=\sum_{i}^{N_{st}}\left(\left[\left[\ell_{1}\left(y_{i}^{s}, y_{i}^{t}\right)\right]\right]\right) \\
&+\frac{\alpha}{N_{st}^{2}}\sum_{i=1}^{N_{st}} \sum_{j=1}^{N_{st}} \left( \left[\left[k\left(h_{i}^{s}, h_{j}^{s}\right)\right]\right]+ \left[\left[k\left(h_{i}^{t}, h_{j}^{t}\right)\right]\right] 
- \left[\left[2k\left(h_{i}^{s}, h_{j}^{t}\right)\right]\right]\right)  \\
&+\left[\left[\frac{\beta}{2} \mathcal{L}_{reg}^{s}\right]\right]+\left[\left[\frac{\beta}{2} \mathcal{L}_{reg}^{t}\right]\right],
\end{aligned}
\end{equation}
where k() denotes the kernel function. By evaluating the derivative of $\theta_{s}$ we obtain:
\begin{equation}
\small
\begin{aligned}
\left[\left[\frac{\partial \mathcal{L}}{\partial \theta^{s}}\right]\right] &=\sum_{i}^{N_{st}}\left(\left[\left[\frac{\partial \ell_{1}\left(y_{i}^{s}, y_{i}^{t}\right)}{\partial \theta^{s}}\right]\right]\right) \\
&+\frac{\alpha}{N_{st}^{2}}\sum_{i=1}^{N_{st}} \sum_{j=1}^{N_{st}}\left(\left[\left[\frac{\partial k\left(h_{i}^{s}, h_{j}^{s}\right)}{\partial \theta^{s}}\right]\right]-2\left[\left[\frac{\partial k\left(h_{i}^{s}, h_{j}^{t}\right)}{\partial \theta^{s}}\right]\right]\right) \\
&+\left[\left[\frac{\beta}{2} \frac{\partial \mathcal{L}_{reg}^{s}}{\partial \theta^{s}}\right]\right]
\end{aligned}
\end{equation}
similarly for $\theta_{t}$ we have:
\begin{equation}
\small
\begin{aligned}
\left[\left[\frac{\partial \mathcal{L}}{\partial \theta^{t}}\right]\right] &=\sum_{i}^{N_{st}}\left(\left[\left[\frac{\partial \ell_{1}\left(y_{i}^{s}, y_{i}^{t}\right)}{\partial \theta^{t}}\right]\right]\right) \\
&+\frac{\alpha}{N_{st}^{2}}\sum_{i=1}^{N_{st}} \sum_{j=1}^{N_{st}}\left(\left[\left[\frac{\partial k\left(h_{i}^{t}, h_{j}^{t}\right)}{\partial \theta^{t}}\right]\right]-2\left[\left[\frac{\partial k\left(h_{i}^{s}, h_{j}^{t}\right)}{\partial \theta^{t}}\right]\right]\right) \\
&+\left[\left[\frac{\beta}{2} \frac{\partial \mathcal{L}_{reg}^{t}}{\partial \theta^{t}}\right]\right]
\end{aligned}
\end{equation}

\section{Experiments}
\subsection{Datasets}
We conduct extensive experiments on multiple mainstream datasets: (1) UCI-Credit-Card dataset: This dataset contains information about default payments, demographic factors, credit data, payment history, and billing for credit card customers in Taiwan from April 2005 to September 2005. The overall dataset contains 30,000 records and 23 features. (2) UCI-Census-Income dataset: this dataset was collected from the 1994 Census bureau database. A set of records was obtained by using some filter conditions. We divide the feature space into two disjoint spaces, i.e., one for source domain and another for target domain to achieve transfer learning. Our model consists of several convolutional layers to extract features followed by a sigmoid layer for classification and we set $L_{1}=1$ and $L_{2}=1$ in our experiment. The neuron numbers of the first $L_{1}$ and latter $L_{2}$ are set to 128 and 64, respectively.

\subsection{Impact of different kernel function}
We choose various kernel functions to analyze their impact on the network transferability.  Three popular kernel functions are tested in our experiments. Specifically, we have:
\begin{itemize}
    \item  Linear kernel: $k(\mathbf{x}, \mathbf{y})=\mathbf{x}^{T} \mathbf{y}$
    \item Polynomial kernel: $k(\mathbf{x}, \mathbf{y})=\left(\mathbf{x}^{T} \mathbf{y}+c\right)^{d}$
    \item Gaussian kernel: $k(\mathbf{x}, \mathbf{y})=\exp \left(-\frac{\|\mathbf{x}-\mathbf{y}\|_{2}^{2}}{2 \sigma^{2}}\right)$
\end{itemize}

We compare the model performance under different kernel function setting and present the result in Table 1, Table 2 and Table 3. As shown in Table 1, the encrypted linear kernel version achieves almost the same accuracy as the non-encrypted version. Compared with the safety of the calculation process, the loss of accuracy is almost negligible. For polynomial kernel we use the version of $c=0,d=2$ and $c=0,d=3$. By increasing d we observe a slight performance loss, because higher degree of the polynomial makes the calculation process more complicated. For the Gaussian kernel, we use its second-order Taylor expansion. Among the aforementioned kernel functions, Gaussian kernel obtains the best performance due to its excellent ability to measure the distance. We also run FTL\cite{liu2018secure} on those two datasets and obtain the auc of 0.712 and 0.837, which proves the superiority of our method.
\renewcommand\tabcolsep{2.8pt}
\renewcommand\arraystretch{1.2}
\begin{table}[]
\caption{Experimental results on the UCI-Credit-Card/UCI-Census-Income dataset using linear kernel}
\begin{tabular}{|c|ccc|ccc|}
\hline
\multirow{2}{*}{} & \multicolumn{3}{c|}{UCI-Credit-Card} & \multicolumn{3}{c|}{UCI-Census-Income} \\ \cline{2-7} 
                  & fscore     & auc      & precision    & fscore     & auc    & precision    \\ \hline
encrypted         & 0.684      & 0.722    & 0.695        &0.781            &0.845        &0.791              \\ \hline
w/o encryption    & 0.693      & 0.730    & 0.698        &0.790            &0.853        &0.799              \\ \hline
source only       & 0.665      & 0.701    & 0.662        &0.754            &0.817        &0.763              \\ \hline
\end{tabular}
\end{table}
\renewcommand\tabcolsep{1.6pt}
\begin{table}[]
\caption{Experimental results on the UCI-Credit-Card/UCI-Census-Income dataset using polynomial kernel}
\begin{tabular}{|c|ccc|ccc|}
\hline
\multirow{2}{*}{}                                                 & \multicolumn{3}{c|}{UCI-Credit-Card} & \multicolumn{3}{c|}{UCI-Census-Income}                                  \\ \cline{2-7} 
                                                                  & fscore     & auc      & precision    & fscore               & auc                  & precision             \\ \hline
encrypted(c=0,d=2)                                                & 0.678      & 0.721    & 0.687        &0.782                      &0.846                      &0.790                       \\ \hline
encrypted(c=0,d=3)                                                & 0.671      & 0.723    & 0.678        &0.783                      &0.845                      &0.791                       \\ \hline
\begin{tabular}[c]{@{}c@{}}w/o encrption\\ (c=0,d=2)\end{tabular} & 0.685      & 0.729    & 0.688        &0.790                      &0.854                      &0.798                       \\ \hline
\begin{tabular}[c]{@{}c@{}}w/o encrption\\ (c=0,d=3)\end{tabular} & 0.674      & 0.728    & 0.679        & 0.789 & 0.855 & 0.797 \\ \hline
source only                                                       & 0.665      & 0.701    & 0.662        & 0.754  &0.817 & 0.763 \\ \hline
\end{tabular}
\end{table}
\renewcommand\tabcolsep{2.2pt}
\begin{table}[]
\caption{Experimental results on the UCI-Credit-Card/UCI-Census-Income dataset using Gaussian kernel}
\begin{tabular}{|c|ccc|ccc|}
\hline
\multirow{2}{*}{}                                             & \multicolumn{3}{c|}{UCI-Credit-Card} & \multicolumn{3}{c|}{UCI-Census-Income}                                  \\ \cline{2-7} 
                                                              & fscore     & auc      & precision    & fscore               & auc                  & precision             \\ \hline
encrypted($\sigma=1$)                                                &0.688            &0.724          &0.696              &0.787                      &0.851                      &0.793                       \\ \hline
encrypted($\sigma=2$)                                                &0.686            &0.725          &0.697              &0.786                      &0.859                      &0.796                       \\ \hline
\begin{tabular}[c]{@{}c@{}}w/o encrption\\ ($\sigma=1$)\end{tabular} &0.702            &0.739          &0.703              &0.795                      &0.868                      &0.799                       \\ \hline
\begin{tabular}[c]{@{}c@{}}w/o encrption\\ ($\sigma=2$)\end{tabular} &0.711            &0.743          &0.709              &0.801  &0.867  &0.808  \\ \hline
source only                                                   & 0.665      & 0.701    & 0.662        & 0.754  &0.817 & 0.763 \\ \hline
\end{tabular}
\end{table}

\subsection{Transfer learning vs source-only}
To verify the effectiveness of transfer learning, we also compare the performance of transfer learning method and source only method. In the source-only version, we only use the source domain data without accessing the target domain data. Compared with the transfer learning method, the performance of the source-only method will be greatly reduced, which also proves the effectiveness of transfer learning.

\section{Conclusion}
In this paper we propose a secure distance measurement metric, named secure maximum mean discrepancy (SMMD), to avoid the potential information leakage in transfer learning. Compared with existing secure transfer learning methods that always suffer from significant performance drop, our proposed MMD is almost as precise as the non-secure version. Our future work includes a more comprehensive analysis of the impact of kernel functions on the transfer performance.

\bibliographystyle{ACM-Reference-Format}
\bibliography{ref}

\end{document}